# Efficient training and design of photonic neural network through neuroevolution

**Tian Zhang[1], Jia Wang[1], Yihang Dan[1], Yuxiang Lanqiu[1], Jian Dai[1], Xu Han[2], Xiaojuan Sun[3] and Kun Xu[1*]**

[1]*State Key Laboratory of Information Photonics and Optical Communications, Beijing University of Posts and Telecommunications, Beijing 100876, China*
[2]*Huawei Technologies Co., Ltd, Shenzhen 518129, Guangdong, China*
[3]*School of Science, Beijing University of Posts and Telecommunications, Beijing 100876, China*
*[\*xukun@bupt.edu.cn](mailto:xukun@bupt.edu.cn)*

**Abstract:** Recently, optical neural networks (ONNs) integrated in photonic chips has received extensive attention because they are expected to implement the same pattern recognition tasks in the electronic platforms with high efficiency and low power consumption. However, the current lack of various learning algorithms to train the ONNs obstructs their further development. In this article, we propose a novel learning strategy based on neuroevolution to design and train the ONNs. Two typical neuroevolution algorithms are used to determine the hyper-parameters of the ONNs and to optimize the weights (phase shifters) in the connections. In order to demonstrate the effectiveness of the training algorithms, the trained ONNs are applied in the classification tasks for iris plants dataset, wine recognition dataset and modulation formats recognition. The calculated results exhibit that the training algorithms based on neuroevolution are competitive with other traditional learning algorithms on both accuracy and stability. Compared with previous works, we introduce an efficient training method for the ONNs and demonstrate their broad application prospects in pattern recognition, reinforcement learning and so on.

## 1. Introduction

Artificial neural networks (ANNs), deep learning [1] in particular, has attracted a great deal of research attentions for an impressively large number of applications, such as image processing [2], natural language processing [3], acoustical signal processing [4], time series processing [5], self-driving [6], games [7], robot [8] and so on. It should be noted that the training of the ANNs with deep hidden layers, especially for convolutional neural networks (CNNs) and recurrent neural networks (RNNs), for example AlexNet [9], VGGNet [10], GoogLeNet [11], ResNet [12] and long short-term memory [13], typically demands significant computational time and resources [1]. Thus, various electronic special-purpose platforms based on graphical processing units (GPUs) [14], field-programmable gate arrays (FPGAs) [15] and application-specific integrated circuits (ASICs) [16] were invented to accelerate the training and inference process of deep learning. On the other hand, in order to obtain general artificial intelligence, some brain-inspired chips including IBM TrueNorth [17], Intel Loihi [18], and SpiNNaker [19] were designed by imitating the structure of a brain. However, even both energy efficiency and speed were improved, the performances of the brain-inspired chips were difficult to compete with the state of the art of deep learning [20]. In the recent years, optical computing had been demonstrated as an effective alternative to traditionally electronic computing architectures and expected to alleviate the bandwidth bottlenecks and power consumption in electronics [21]. For example, new photonic approaches for spiking neuron and scalable network architecture based upon excitable lasers, broadcast-and-weight protocol and reservoir computing had been illustrated [22-24]. Despite ultrafast spiking response were achieved, these neuromorphic photonic systems also faced the similar challenge of performance and integration issues [24]. Y. Shen *et al.* proposed a photonic implementation of ANNs and pointed out this integrated optical neural networks (ONNs) architecture could improve the computational speed and consumption compared with the conventional computers [25]. In addition, the photonic chips that could implement the similar functions of CNNs [26] and RNNs [27] were outlined based on the ONNs. X. Lin *et al.* introduced an all-optical diffractive deep neural network that could complete feature detection and image recognition [28]. Besides, several different nonlinear activation functions based on electronic and photonic elements were proposed in the ONNs to enhance the nonlinear capability [29, 30]. It should be noted that the back-propagation (BP) and stochastic gradient descent (SGD) were currently used as the training methods for the ONNs [25]. However, the BP and SGD training strategies were difficult to implement in the integrated optical chips, thus the determination of the weights in the ONNs were generally mapped from the pre-trained results on a digital computer [25]. Obviously, this train method was inefficient because of the restricted accuracy of the model representation and the loss of the advantages in speed and energy [31].

In order to self-learn the weights in the ONNs, *in situ* computation of the gradient for weights based on brute force had been reported [25]. Similarly, a self-learning photonic signal processor trained by a modified SGD was proven experimentally to implement tunable filter, optical switching and descrambler [32]. T.W. Hughes *et al.* found that the brute force method had high computational complexity for large systems. Thus, they proposed a novel training method to compute the gradients of weights by using the *in situ* intensity measurements and adjoint variable method (AVM) [33]. Here, the determinations of the gradients in the ONNs were converted to an inverse design and sensitivity analysis process for photonic circuits [33]. In addition, some training strategies and tricks used in deep learning, such as adaptive moment estimation and initialization scheme were also applied in training of the ONNs [29, 31]. The training methods proposed by T.W. Hughes inspire us that the learning process of the ONNs can be converted to the inverse design problems that are solved by using gradient-based methods or gradient free methods [34, 35]. Apart from gradient-based methods (such as the AVM), gradient free methods, for example genetic algorithms (GA) and particle swarm optimization (PSO), also can be applied in the inverse design of photonic devices [36-39]. Besides that, as an alternative approach to train ANNs, neuroevolution, which derives from the evolution process imitated the biological brain, is a typical gradient free method based on evolutionary algorithms [40]. Compared with gradient-based methods, such as the SGD and AVM, neuroevolution can not only determine the weights in the connections but also optimize the network architectures of ANNs, hyper-parameters of activation functions and the rules for learning algorithm owing to the advantages of diversity, parallelization and architecture search [40]. More interestingly, it has been proved that neuroevolution performs competitively with the learning algorithms in the deep reinforcement training (DRT), such as policy search and deep Q-network [41]. On a subset of Atari games, neuroevolution even outperforms the best training algorithms in DRT on the training speed because of its parallelizable [42]. Obviously, if the ONNs can be trained by neuroevolution, it not only provides a novel learning method for the ONNs but also is expected to improve the training efficiency for the traditional DRT.

In this article, we propose a novel learning strategy to design and train the ONNs based on neuroevolution. Two typically evolutionary algorithms, GA and PSO are used to determine the hyper-parameters of the ONNs and optimize for the weights (phase shifters) in the connections. In order to demonstrate the effectiveness of neuroevolution, the trained ONNs are applied in the classification tasks for different datasets. The calculated results exhibit that these simple training algorithms are competitive with traditional learning algorithms (such as SGD and AVM) and have broad application prospects in pattern recognition and DRT.

## 2. Training methods based on neuroevolution

As shown in Fig. 1(a), the network architecture of ANNs imitates the structure of biological neural network which includes a great number of neurons and connections layer by layer [1]. It should noticed that although ANNs are brain-inspired, there are significant differences in the network structure, learning method, information transmission and encoding rule compared with biological brain [24]. For example, the information propagated between the biological neurons is conveyed by synapses and the code scheme of it is spike timing [24]. Owing to the temporal coding and event-driven manner, spiking neural networks (SNNs) are the closest approximation to real neural network that has computational advantages in energy efficiency and high speed [43]. Nevertheless, although SNNs have been applied in many areas (such as image and speech recognition [44, 45]), actual performance and application range are difficult to compete with deep learning. In order to combine the advantages of ANNs and optical computing, as shown in Fig. 1(b), a photonic implementation of ANNs that includes optical interference unit (OIU) and optical nonlinearity unit (ONU) has been outlined to complete the speech recognition task experimentally [25]. From Fig. 1(b), it can be found that the network architecture of the ONNs strictly imitates that of the ANNs, namely, the OIU and ONU implement matrix multiplication and nonlinear transform functions, respectively [25]. As shown in Fig. 1(c), the physical implementation of the OIUs in the ONNs are composed of programmable Mach–Zehnder interferometers (MZIs) whose phase shifters are controlled by external voltage to construct any unitary matrix [25]. While the ONUs in the ONNs can be realized by means of the strong nonlinear effects of two dimension (2D) materials, such as graphene and sulfide [46]. However, the integrated fabrication of 2D materials into the silicon waveguide is complicated and the strength of nonlinear effect is weaken when the 2D materials are integrated in the waveguide [31]. In order to alleviate the shortcomings of optical nonlinearity integrated in the waveguide, an electro-optic hardware platform which implements various nonlinear activation functions with low activation threshold is illustrated in Ref.[31].

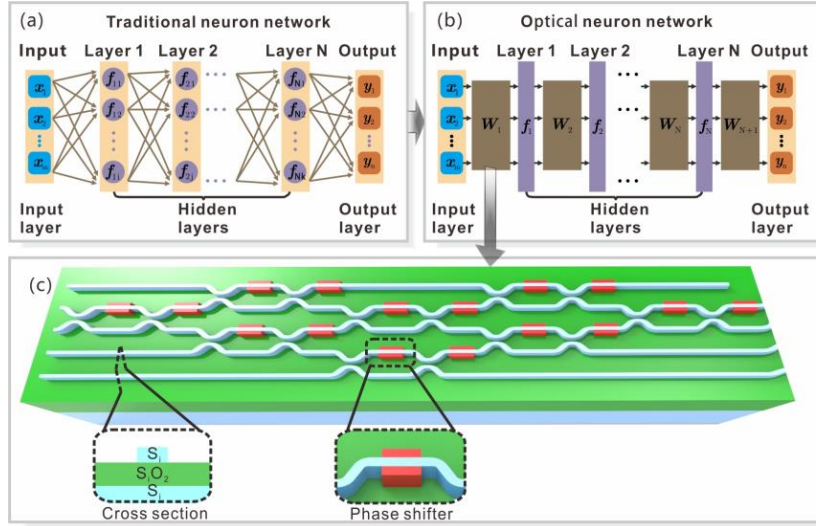

Fig. 1. (a) The network architecture of the ANNs which includes input neurons, hidden layers and output neurons. (b) Decomposition of the ANNs into a series of layers which implement linear matrix multiplication and nonlinear transform functions. (c) Physical implementation of the OIU that are composed of programmable MZIs.

Training the neural network based on optimization algorithms is a critical requirement and procedure regardless of ANNs and ONNs [33]. The BP and SGD methods are representative gradient-based optimization methods to compute the gradients of the model parameters based on the chain rule [33]. The derivative of the loss backwards from the output layer to the input layer of ANNs. The update rule of model parameters is given by

$$\omega \leftarrow \omega - \eta(\alpha \frac{\partial R(\omega)}{\partial \omega} + \frac{\partial L}{\partial \omega}) \qquad (1)$$

where $\eta$ is the learning rate, $L$ is a loss function that measures model performance, $R$ is a regularization term and $\alpha$ is penalty, respectively [33]. As same as ANNs, the ONNs in Ref. [25] are trained by using the BP and SGD algorithms. As alternative approaches to gradient-based methods, evolutionary algorithms are representative gradient free methods to optimize the weights of ANNs [41, 42]. Through the selection, crossover and mutation processes of a population, evolutionary algorithms provide a natural choice way to gradually optimize the model parameters to achieve a superior fitness. There are many attractive reasons for using evolutionary algorithms to train ANNs, the most important one is that the evolutionary algorithms optimize more hyper-parameters that are not considered in the SGD. For example, the neuroevolution of augmenting topologies (NEAT) algorithm is a typical search method which solves the problem of crossing over variable network topologies through historical marking [40]. In this article, neuroevolution is proposed to determine the hyper-parameters of the ONNs and optimize for the weights in the connections. The network architectures of the ONNs are modelled by a chip-level simulation platform, neuroptica [47]. Neuroptica provides a range of abstraction levels encapsulated by Keras-like application programming interfaces (APIs) [47]. The lowest abstraction levels and the highest abstraction levels in the neuroptica manipulate the properties of phase shifters and whole network architectures of the ONNs, respectively [47]. It has been demonstrated that the functions of an logical gate can be effectively implemented by using the neuroptica based on the BP and SGD training strategies [31]. In addition, neuroptica provides two decomposition ways (Reck [48] and Clements [49] decompositions) to construct any unitary matrices by arranging the phase shifters and MZIs in the optical meshes reasonably. While the optical neurons in the ONNs implement nonlinear activation functions based on the optical-to-optical nonlinearities which have advantages in expressiveness compared with the all-optical nonlinearities [31]. The nonlinear activation function can be fabricated in an electro-optic hardware platform which is configured by three critical physical parameters: amount of power tapped off to photodetector $\alpha$, phase gain $g$ and the biasing phase $\theta$ [31]. We will included the three physical parameters as optimized variables in the training algorithms except for the weights between the optical neurons. Moreover, two kinds of traditional optimizers in the neuroptica can be used to train the ONNs based on the AVM and SGD (or adaptive moment estimation) [31].

Two kinds of typical gradient free algorithms, the GA and PSO, are tried to validate the effectiveness of the training algorithms based on neuroevolution. The GA is a representative neuroevolution algorithm that is widely used in inverse design and performance optimization of photonic devices [37, 38]. In this article, we use the GA to train the ONNs by optimizing for the phase shifters of optical mesh and hyper-parameters of nonlinear activation functions. The algorithmic details of the GA are outlined as follows: (i) selecting a reasonable network architecture of the ONNs for targeted dataset. Then randomly generating a population of $N$ individuals (ONNs) that have the same network architectures but different weights and hyper-parameters. For example, we construct a simple network architecture consisted of $L$=3 layers for the famous iris plants dataset. In the iris plants dataset, the features of each iris plant and the categories of all iris plants are 4 and 3, respectively. As a result, the input layers of the ONNs includes 4 input ports, while the output layers includes 3 output ports. Each hidden layer in the ONNs includes a 4×4 unitary matrix implemented by an optical mesh and a layer of 4 parallel electro-optic activation functions proposed in [31]. In the final layer, the electro-optic activation functions are replaced by a detection layer which corresponds to the power measurement by using a photodetector [31]. After the final layer, a dropmask layer is added in the network to reduce the dimension from 4 to 3 by abandoning a output port. Here, the dropmask layer can be added in the hidden layer to make the dimension of the intermediate layer more flexible (larger or smaller than 4). In order to implement arbitrary unitary matrix, the unitary optical mesh includes a layer of phase shifters at the beginning of the optical mesh [47]. The optical responses of electro-optic activation functions are configured by three hyper-parameters $\alpha$, $g$ and $\theta$ which are also considered in the GA. For all the generated ONNs, all optimization variables are initialized in the different ranges specified by minimum and maximum values $0<\varphi<2\pi$, $0<\alpha<1$, $0<g<\pi$ and $-2\pi<\theta<0$, where $\varphi$ is the phase of phase shifter. (ii) For the generated ONNs, training instances are input into the ONNs and transfers in the ONNs from input layer to output layer. This process is called as the forward-propagation step where the training instances are performed by linear operations (unitary optical mesh) and nonlinear operations (electro-optic activation function) layer by layer. In the output layer, the predicted results for current iteration are collected, and the prediction losses between the prediction results and target results are calculated based on the categorical cross entropy or mean squared error (MSE) methods. Here, the prediction losses for all the generated *N* ONNs are calculated and sorted in ascending order. (iii) Trying to generate a new population of the ONNs by using the standard selection, cross and mutation procedures. In the selection process, the prediction loss of the ONNs are regarded as fitness. Two parent individuals are selected from the previous generation based on the roulette-wheel selection method or tournament strategy [36]. The ONNs with smaller prediction loss are selected with the higher probability in the selection process. In order to maintain the diversity of population or keep some superior individuals, some percentage of the superior ONNs or inferior ONNs are retained in the next population. In the crossover process, the optimization variables consisted of weights and hyper-parameters are extracted from the ONNs and converted into the binary values. It should be noted that the conversion of decimal number to binary number is likely to result in the loss of digital precision. The optimization variables of the parent individuals (ONNs) cross over to generate a new individual based on the uniform crossover algorithm [36]. Here, in the uniform crossover algorithm, the probabilities of gene exchange and crossover are 0.5 and 0.8, respectively. In the mutation process, each element in the binary number has 5% probability to flip from 0 (1) to 1 (0). After converting the weights and hyper-parameters from binary to decimal, new individuals (ONNs) are generated. The new generated ONNs and the remaining ONNs selected from the previous generation form a new population. (iv) Evaluating the performance metrics of the generated population and determining the training process whether stop or not. In our training algorithm, if the generation of the ONNs evolves for 1000 times or the prediction losses remain unchanged for more than 5 generations, the training process stops, otherwise, proceeds to Step (ii). Flowchart of the learning process for the ONNs based on the GA is shown in Fig. 2(a).

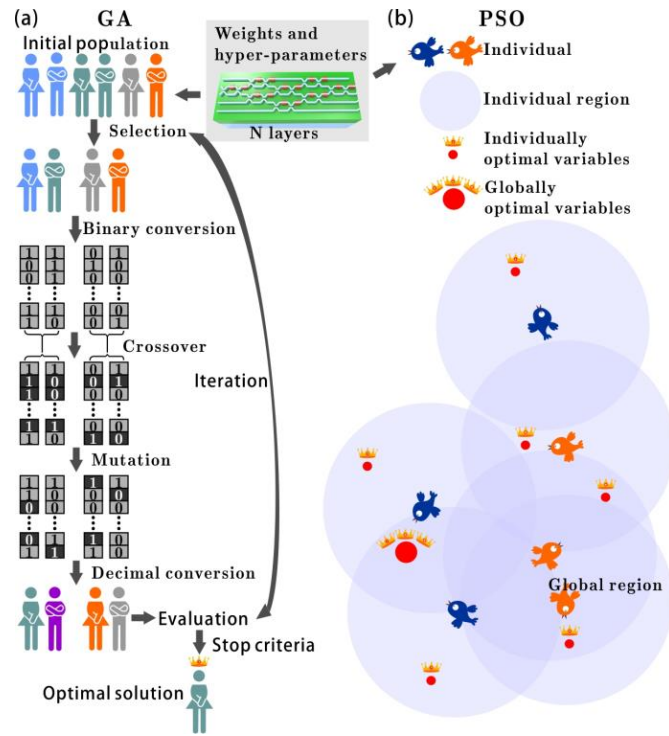

Fig. 2. Flowcharts of the learning algorithms for the ONNs based on the GA (a) and PSO (b).

Similar to the GA, the PSO is an evolutionary algorithm which is suitable for the decimal number rather than binary number [39]. The generation of initial population for the PSO is the same as that of the GA. However, the generation of new population for the PSO is not through selection, crossover and mutation procedures. It indicates that there no need to convert the decimal number to binary number in the PSO which can effectively avoid the loss digital precision. Flowchart of the learning process for the ONNs based on the PSO is shown in Fig. 2(b). For the PSO, the individuals in the population depend on the currently optimal individual and historically optimal solution to evolutionally optimize for the fitness [39]. Similarly, when we use the PSO to train the ONNs, each ONN in the population searches for the optimal variables (weights and hyper-parameters) by synthetically considering the globally optimal variables and individually optimal variables. It indicates that the evolution of the optimization variables for all ONNs are controlled by a specified velocity [50]:

$$V_i^{k+1} = WV_i^k + c_1 r_1 \left( pb_i^k - X_i^k \right) + c_2 r_2 \left( gb_k^d - X_i^k \right) \qquad (2)$$

where $i$ represents the $i$th ONN in the population, $k$ is iteration number, $W$ is inertia weight, $c_1=c_2=1.49445$ are acceleration constants, $r_1$ ($r_2$) is random values between 0 and 1, $gb_k^d$ relates to the globally optimal weights and hyper-parameters for all ONNs, $X_{ik}$ and $pb_i^k$ are the current variables and individually optimal variables for the $i$th ONN in the $k$th iteration, respectively. And the weights and hyper-parameters of the $i$th ONN are updated according to following equation [50]:

$$X_i^{k+1} = X_i^k + V_i^{k+1} \qquad (3)$$

In order to avoid the premature problem, the velocities of evolution are limited to a certain range (-2~2). Finally, in each iteration, the prediction losses of the newly generated population are evaluated to determine the training process whether stop or not. If the generation evolves for 1000 times or the prediction losses remain unchanged for more than 5 generations, then the PSO stops.

## 3. Calculated results and discussions

As typical datasets in the classification tasks, the iris plants dataset [51] and wine recognition dataset [52] are selected as the test datasets to demonstrate the effectiveness of the training algorithm. The iris plants dataset is a simple dataset which includes 150 instances (4 attributes for each instance). Compared with the iris plants dataset, wine dataset is more complex due to the 13 features for each instance, leading to demand a more complex network architecture of the ONNs. We should construct a ONN which includes 13 input ports and 3 output ports to recognize the wine dataset. In addition, to verify the practical applications of the training algorithm based on neuroevolution, we use the ONNs to recognize the modulation formats in a communication system. Here, we don't utilize the complete digital signals to train the ONNs because of their high dimensions. Alternatively, four effective attributes $\gamma_{max}$, $\sigma_{aa}$, $\sigma_{dp}$ and $\sigma_{af}$ are extracted from 800 digital signals randomly modulated by 4ASK, 4FSK, BPSK and QPSK modulation formats [53]. It has been demonstrated that these statistical features can exhibit obvious separation in distributions for those modulation formats [53]. Here, the four effective features are inputted into the ONNs which are trained by supervised learning to identify the maximum possible category of the modulation formats. Besides, we also use two kinds of randomly generated datasets provided by the neuroptica simulation platform to compare the training effects between the evolutionary algorithms (GA and PSO) and AVM. As shown in Fig. 3(a) and Fig. 4(a), the data points in the example datasets are randomly generated to segment specific square areas into triangle (ring) part and other part [47]. It indicates that the recognition of the example datasets is a binary classification problem whose categories are coded by one-hot labelling. We use 80 percent (320) and 20 percent (80) of all the instances (400) in the example datasets as the training set and test set, respectively.

Fig. 3 exhibits the calculated results of the ONNs trained by the GA. First of all, we use the APIs provided by the neuroptica simulation platform to generate a simple dataset which segments the square space into two triangle areas. As shown in Fig. 3(a), the red cross marks and blue circles correspond to two triangle areas respectively. The network architectures of the ONNs includes $L$=5 layers decomposed by Clements methods [49] and each layer includes 5 neurons. Similar to the demo provided by the neuroptica simulation platform, we reshape the input data to fit into the specified mesh size and normalize them to have the same total power.

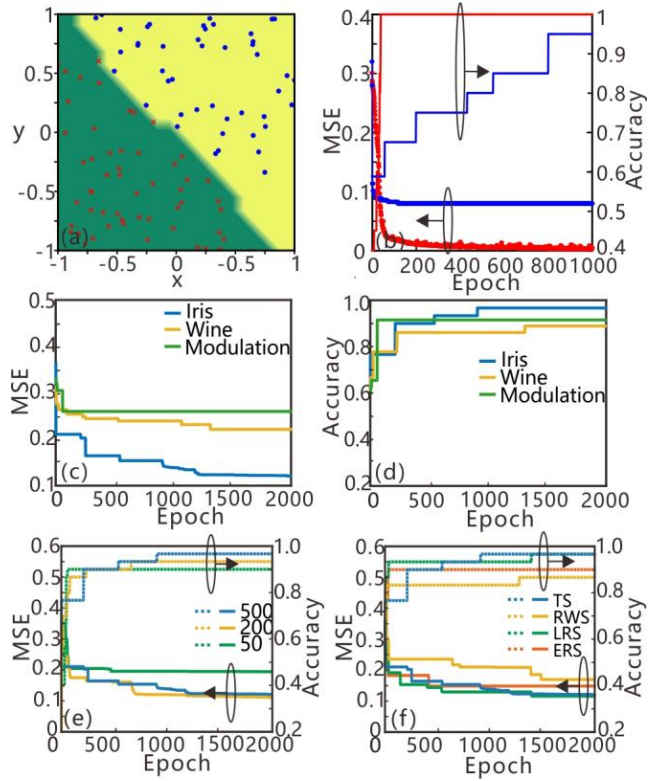

Fig. 3. The calculated results of the ONNs trained by the GA. (a) A simple dataset provided by the neuroptica simulation platform segments the square space into two triangle areas. (b) The MSEs and classification accuracies of the ONNs trained by the AVM and GA. (c) The MSEs of the ONNs trained by the GA for three test datasets. (d) The classification accuracies of the ONNs trained by the GA for three test datasets. The MSEs and classification accuracies of the trained ONNs with different population sizes (e) and selection operators (f) of the GA.

The ONNs are trained by the GA and AVM, and the calculated results are exhibited in Fig. 3(b). Here, the population size of the GA is set as $N$=500, and the MSEs between the prediction results and the target results are chosen as the fitness for the GA. It can be found in Fig. 3(b) that the prediction losses of the AVM (red circles) decrease from 0.35 to 0.005, while that of the GA (blue circles) reduce from 0.38 to 0.09, suggesting the learning algorithms are convergent for the training datasets. Although both the GA and AVM are effective for the simple pattern recognition, the performance of GA is worse than that of the AVM. And this conclusion can be verified by the variations of accuracies (solid line) for the GA and AVM in Fig. 3(b). It can be observed that the accuracies of the AVM and GA can reach to 1 and 0.95, respectively. And the AVM can achieve an excellent accuracy (>0.95) faster than the GA. Obviously, although the GA training method can't compete with the AVM learning method on the accuracy and speed, it is still effective for the training of the ONNs. The contours shown in Fig. 3(b) is the classification boundary for the GA. Except for the simple example dataset, we also use the ONNs trained by the GA to recognize three practical dataset, namely, the iris plants dataset, wine dataset and modulation format recognition dataset. For particular network architectures consisted of $L$=3 layers for the iris plants dataset (modulation format recognition dataset) and $L$=2 layers for the wine recognition dataset, the variations of the prediction loss and classification accuracy are illustrated in Fig. 3(c) and Fig. 3(d), respectively. For the 3 layer ONNs and 2 layer ONNs, the totals of the optimization variables includes weights and hyper-parameters are 54 and 35, respectively. As shown in Fig. 3(c), the MSEs of the three datasets decrease to 0.12 (iris plants dataset), 0.22 (wine dataset) and 0.26 (modulation format recognition dataset) after 2000 iterations, which indicates the convergence of the GA. From

Fig. 3(d), the classification accuracies for the test datasets reach 0.97 (iris plants dataset), 0.89 (wine dataset) and 0.92 (modulation format recognition dataset), respectively. In fact, the trained algorithm has converged in the vicinity of 100 iterations (the classification accuracies for all datasets are greater than 0.80), which means that the GA is effective for training the ONNs in practical classification applications. In addition, we compare the training effects under different parameters of the GA, such as population size and selection operator. Here, for simplicity, we take the iris plants dataset as an example. Fig. 3(e) exhibits the influences of different population sizes $N$=50, 200 and 500 on the MSE and classification accuracy. Obviously, the larger the population sizes are, the lower the final values of MSE become (0.11 for $N$=500, 0.12 for $N$=200 and 0.20 for $N$=50) and the higher the classification accuracies change into (0.97 for $N$=500, 0.93 for $N$=200 and 0.9 for $N$=50). This phenomenon is easy to understand that the large population size enhance the global searching ability of the GA. Moreover, we also compare the results of different selection operators (such as tournament strategy (TS), roulette-wheel selection (RWS), linear ranking selection (LRS) and exponential ranking selection (ERS)) on the training effects. As shown in Fig. 3(f), we can observe that the classification accuracies of the TS, LRS, RWS and ERS reach to 0.97, 0.97, 0.87 and 0.90, respectively. It indicates that TS and LRS are more effective compared with the other two selection operators.

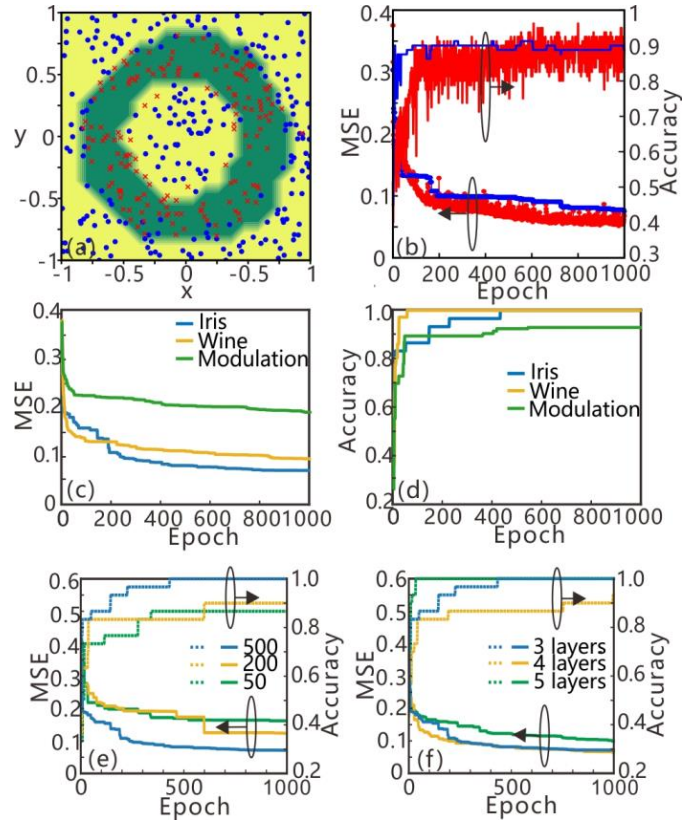

Fig. 4. The calculated results of the ONNs trained by the PSO. (a) A simple dataset provided by the neuroptica simulation platform segments the square space into ring part and other part. (b) The MSEs and classification accuracies of the ONNs trained by the PSO and AVM. (c) The MSEs of the ONNs trained by the PSO for three test datasets. (d) The classification accuracies of the ONNs by the PSO for three test datasets. The MSEs and classification accuracies of the trained ONNs for different population sizes of the PSO (e) and layers of the ONNs (f).

The calculated results of the ONNs trained by the PSO are shown in Fig. 4. Firstly, we use the function provide by the neuroptica simulation platform to generate a simple dataset which segments the square space into ring part and other part. As shown in Fig. 4(a), the red cross marks relate to the ring area, while the blue circles correspond to the other area. The ONNs which consists of *L*=5 layers decomposed by Clements methods [49] (each layer includes 5 neurons) are trained by the PSO and AVM, and the calculated results are exhibited in Fig. 4(b). Here, the prediction losses which are chosen as the fitness for the PSO are calculated based on MSE in each iteration. It can be found in Fig. 4(b) that the prediction losses (red and blue circles) of the AVM and PSO decrease from 0.35 (0.37) to 0.05 (0.07), suggesting that the learning algorithms are convergent for the training datasets. It is noteworthy that the final prediction loss of the PSO is approximation to that of the AVM. This phenomenon can be confirmed by the classification accuracies (red and blue solid line) illustrated in Fig. 4(b). We observe that the classification accuracies of the PSO achieve an excellent accuracy (>0.85) faster than the AVM although both the AVM and PSO can both reach to 0.9 finally. And the contours shown in Fig. 4(b) are the classification boundary for the PSO. Obviously, the PSO training method are competitive with the AVM learning algorithms on accuracy and stability. As same as the GA, we also use the trained ONNs to recognize the iris plants dataset, wine dataset and modulation format dataset. Here, the network architecture of the ONNs contains 3 layers, and each layer includes an optical mesh followed by an electro-optic activation function with intensity modulation [31]. It can be observed from Fig. 4(c) that when the population of the PSO evolves 1000 generations, the MSEs between the prediction results and the target results exhibit tendencies toward to low values, indicating the convergence of the training algorithm. Correspondingly, we observe in Fig. 4(d) that the classification accuracies for the iris plants dataset and wine dataset reach to 100%. For the relatively complex dataset (modulation format dataset), the classification accuracy also increases to 0.93. Obviously, the accuracies of the ONNs trained by the PSO are superior to those of the ONNs trained by the GA. The reason for this phenomenon is attributed to that there is no need to convert the optimization variables from decimal to binary for the PSO, which avoids the loss of digital precision efficiently. And the optimization strategy based on the globally optimal variables and individually optimal variables accelerates the convergence of the training algorithm. Similar to the GA, we also compare the training results under different population sizes and network architectures for the PSO. Here, we also utilize the iris plants dataset to evaluate the training effects. In Fig. 4(e), it can be observed that the classification accuracies can reach to 1.0, 0.9 and 0.87 when the sizes of population for the PSO are set as $N$=500, 200 and 50, respectively. And the reductions of the prediction losses for the large populations ($N$=500 and 200) are superior to that for the small population ($N$=50). This phenomenon is easy to explain because the large populations enhance the global searching ability of the evolution algorithms [36]. Although we can increase the population size of the PSO to achieve the lower MSE value and higher classification accuracy, it's at the expense of the training time. In addition, we also consider the influences of the ONNs with different network architectures on the performance metrics. As shown in Fig. 4(f), the network architectures of the ONNs have not a linear relationship with the MSE and classification accuracy. Here, it can be found that the simplest ONN with 3 layers has a low MSE (0.08) and a high accuracy (100%) in the last iteration. The complex network architecture of the ONN will lead to the over-fitting problem which reduce the accuracy on the test dataset. As a result, the performance of the 5-layer ONN is worse than that of the 3-layer ONN in Fig. 4(f).

## 4. Conclusions

In conclusion, we propose a novel learning strategy to design and train the ONNs based on neuroevolution. Two typical gradient free algorithms, GA and PSO are used to determine the hyper-parameters of the ONNs and optimize the weights (phase shifters) in the connections. To demonstrate the effectiveness of the training algorithms, the trained ONNs are applied in the classification tasks for different datasets. The calculated results exhibit that these simple training algorithms are competitive with other traditional learning algorithms. Compared with previous works, we introduce an efficient training method for the ONNs and demonstrate their broad application prospects in pattern recognition and DTR.

**Funding**

This work was supported by National Natural Science Foundation of China (Grant No. 61625104, No. 61431003); The Beijing Municipal Science & Technology Commission (Grant No. Z181100008918011); The Fundamental Research Funds for the Central Universities (Grant No. 2019RC15, No. 2018XKJC02); National Key Research and Development program (Grant No.2016YFA0301300).